\documentclass[letterpaper]{article}
\usepackage{aaai}
\usepackage{times}
\usepackage{helvet}
\usepackage{courier}
\usepackage{amsmath}
\usepackage[]{subfigure}
\usepackage{array,multirow,graphicx}
\usepackage{cite}
\usepackage[numbers]{natbib}
\usepackage{amssymb}
\usepackage{url}
\usepackage{hyperref}
\usepackage[switch]{lineno}

\usepackage{xcolor}

\begin{document}
%
\title{Deep Probabilistic Imaging: Uncertainty Quantification and \\ Multi-modal Solution Characterization for Computational Imaging}
\author{He Sun, Katherine L. Bouman\\
California Institute of Technology, 
1200 E California Blvd, Pasadena, California 91125
}
\maketitle
\begin{abstract}
\begin{quote}
Computational image reconstruction algorithms generally produce a single image without any measure of uncertainty or confidence. 
Regularized Maximum Likelihood (RML) and feed-forward deep learning approaches for inverse problems typically focus on recovering a point estimate. 
This is a serious limitation when working with underdetermined imaging systems, where it is conceivable that multiple image modes would be consistent with the measured data. Characterizing the space of probable images that explain the observational data is therefore crucial. 
In this paper, we propose a variational deep probabilistic imaging approach to quantify reconstruction uncertainty. Deep Probabilistic Imaging (DPI) employs an untrained deep generative model to estimate a posterior distribution of an unobserved image. This approach does not require any training data; instead, it optimizes the weights of a neural network to generate image samples that fit a particular measurement dataset.
Once the network weights have been learned, the posterior distribution can be efficiently sampled. 
We demonstrate this approach in the context of interferometric radio imaging, which is used for black hole imaging with the Event Horizon Telescope, and compressed sensing Magnetic Resonance Imaging (MRI).
\end{quote}
\end{abstract}

\vspace{-0.0in}
\section{Introduction} \label{sec:intro}

\noindent Uncertainty quantification and multi-modal solution characterization are essential for analyzing the results of underdetermined imaging systems. In computational imaging, reconstruction methods solve an inverse problem to recover a hidden image from measured data. 
When this inverse problem is ill-posed 
there are infinite image solutions that fit the observed data. 
Occasionally these varied images lead to different scientific interpretations; thus it is important to be able to characterize the distribution of possibilities.
In a Bayesian framework, this problem could ideally be addressed by accurately modeling the measurement noise, formulating an estimation problem, and computing the posterior distribution of the hidden image. However, this analytical approach is only tractable in simple cases. When the inverse problem is non-convex or the measurement noise is complicated (e.g., non-Gaussian) the posterior distribution can quickly become intractable to compute analytically.

In this paper, we propose \textit{Deep Probabilistic Imaging} (DPI), a new variational inference approach that employs a deep generative model to learn a reconstructed image's {\it posterior distribution}. More specifically, we represent the posterior probability distribution using an invertible flow-based generative model. By training with a traditional maximum {\it a posteriori} (MAP) loss, along with a loss that encourages distribution entropy, the network converges to a generative model that approximately samples from the image posterior; the model returns a posterior sample image as output for a random Gaussian sample input.
Our proposed approach enables uncertainty quantification and multi-modal solution characterization in  non-convex inverse imaging problems. 
We demonstrate our method on the applications of compressed  sensing MRI and astronomical radio interferometric imaging. High resolution radio interferometric imaging often requires a highly non-convex forward model, occasionally leading to multi-modal solutions.
 



\section{Related Work} \label{sec:related}
\subsection{Computational Imaging}

The goal of a computational image reconstruction method is to recover a hidden image from measured data. Imaging systems are often represented by a deterministic forward model, $y = f(x)$, where $y$ are the observed measurements of a hidden image, $x$. A regularized maximum likelihood (RML) optimization can be solved to reconstruct the image:
\begin{equation} \label{eq:rml}
    \hat{x} = \textrm{arg} \min_x \{ \mathcal{L}(y, f(x)) + \lambda \mathcal{R}(x) \},
\end{equation}
where $\hat{x}$ is the reconstructed image, $\mathcal{L}(\cdot, \cdot)$ is the data fitting loss between the measured data and the forward model prediction, $\mathcal{R}(\cdot)$ is a regularization function on the image, and $\lambda$ is the hyper-parameter balancing the data fitting loss and the regularization term. The regularization function is necessary for obtaining a unique solution in under-constrained systems. Commonly used regularization functions include L1 norm, total variation (TV) or total squared variation (TSV)~\cite{bouman1993generalized, kuramochi2018superresolution}, maximum entropy (MEM)\cite{skilling1984maximum}, multivariate Gaussian~\cite{zoran2011learning}, and sparsity in the wavelet transformed domain~\cite{candes2007sparsity}. 

Assuming the forward model and measurement noise statistics are known, one can derive the probability of measurements, $y$, conditioned on the hidden image, $x$, as $p(y|x)$. In a Bayesian estimation framework, the regularized inverse problem can be interpreted as a maximum a posteriori (MAP) estimation problem:
\begin{equation} \label{eq:map}
    \hat{x} = \textrm{arg} \max_x \{ \log p(y|x) +  \log p(x) \},
\end{equation}
where the log-likelihood of the measurements correspond to the negative data fitting loss in Eq.~\ref{eq:rml}, and the prior distribution of the image defines a regularization function (also referred to as an image prior).

\subsubsection{Deep Image Prior}
In lieu of constraining the image via an explicit regularization function, $\mathcal{R}(\cdot)$, Deep Image Prior approaches\cite{ulyanov2018deep} parameterize the image as a deep neural network and implicitly force the image to respect low-level image statistics via the structure of a neural network:
\begin{equation} \label{eq:deepprior}
    w^{\star} = \textrm{arg} \min_w \mathcal{L}(y, f(g_w(z))) \hspace{0.05in} \mbox{   where  } \hspace{0.05in} \hat{x} = g_{w^{\star}}(z).
\end{equation}
In this approach, $g_w(\cdot)$ is a deep generator, $w$ are the neural network weights, $w^{\star}$ are the optimized weights, and $z$ is a randomly chosen, but fixed, hidden state. The network weights are randomly initialized and optimized using gradient descent. This approach has been demonstrated in many applications, including inpainting, denoising and phase retrieval\cite{ulyanov2018deep,bostan2020deep}. 
Note that since the hidden state, $z$, is fixed, only a single reconstructed image is derived after optimization.

\subsection{Uncertainty Quantification}
Uncertainty quantification is important for understanding the confidence of recovered solutions in inverse problems. 

\subsubsection{MCMC Sampling}
One widely used approach for quantifying the uncertainty in image reconstruction is the Markov Chain Monte Carlo (MCMC) method. 
MCMC provides a way to approximate the posterior distribution of a hidden image via sampling~\cite{bardsley2012mcmc, broderick2020hybrid}. 
However, the MCMC approach can be prohibitively slow for high dimensional inverse problems~\cite{cai2018uncertainty}.

\subsubsection{Bayesian Hypothesis Testing}
Uncertainty quantification can also be formulated as a Bayesian hypothesis test in linear imaging inverse problems \cite{repetti2019scalable}. In a linear imaging problem, the data likelihood is a log-concave function, which makes the hypothesis test on a specific image structure an efficient convex optimization program. However, this method cannot be generalized to imaging inverse problems, when either the negative data likelihood or the regularizer is non-convex.

\subsubsection{Variational Bayesian Methods}
Variational inference is widely used to approximate intractable posterior distributions~\cite{blei2017variational, 2019Arras}. Instead of directly computing the exact posterior distribution, variational Bayesian methods posit a simple family of density functions and solve an optimization problem to find a member of this family closest to the target posterior distribution. Variational Bayesian methods are much faster than sampling methods (e.g., MCMC), and typically achieve comparable performance~\cite{gershman2012nonparametric}. The performance of variational methods depends on the modeling capacity of the density function family and the complexity of the target distribution. A commonly used technique for simplifying the density function is the mean-field approximation\cite{jordan1999introduction}, where the distributions of each hidden variables are assumed independent. The density function can also be parameterized using neural networks, such as flow-based generative models~\cite{rezende2015variational}. 


\subsubsection{Bayesian Neural Networks}
Deep learning has become a powerful tool for computational imaging reconstruction\cite{ongie2020deep}. 
Most current deep learning approaches only focus on the point estimate of the hidden image; however, 
 Bayesian neural networks \cite{mackay1995bayesian,gal2016uncertainty} have been developed to quantify the uncertainty of deep learning image reconstructions\cite{xue2019reliable}. 
In particular, the weights of a Bayesian network are modeled as probability distributions, so that different predictions are obtained every time the network is executed. Although these approaches can achieve impressive performance, they rely on supervised learning and only describe the reconstruction uncertainty conditioned on a training set. 


\subsubsection{Empirical Sampling}
An alternative empirical approach for obtaining multiple candidate reconstructions is to solve a regularized inverse problem (Eq.~\ref{eq:rml}) multiple times with different choices of regularizer hyper-parameters (e.g. $\lambda$ in Eq.~\ref{eq:rml}) and image initializations. This approach was used in~\cite{akiyama2019first} to characterize the uncertainty of the reconstructed black hole image M87*. Although the mean and standard deviation of these images provide a measure of uncertainty, there is no expectation that these samples satisfy a posterior distribution defined by the measurement data. In fact, this method only quantifies the reconstruction uncertainty due to choices in the reconstruction methods, such as regularizer hyper-parameters, instead of the uncertainty due to measurement noise and sparsity.



\subsection{Flow-based Generative Models} \label{sec:flow}
Flow-based generative models are a class of generative models used in machine learning and computer vision for probability density estimation. These models approximate an arbitrary probability distribution by learning a invertible transformation of a generic distribution $\pi(z)$ (e.g. normal distribution). Mathematically, a flow-based generative model is represented as
\begin{equation}
    x = G_{\theta}(z), \quad z = G_{\theta}^{-1}(x),
\end{equation}
where $G_{\theta}(\cdot)$ is an invertible deep neural network parameterized by $\theta$ that links a sample from the target distribution, $x$, with a hidden state, $z$. The application of invertible transformations enables both efficient sampling, as well as exact log-likelihood computation. 
The log-likelihood of a sample can be analytically computed based on the ``change of variables theorem":
\begin{equation} \label{eq:changeofvar}
    \log q_{\theta}(x) = \log \pi(z) - \log \left| \det \frac{d G_{\theta}(z)}{d z} \right|,
\end{equation}
where $\det \frac{d G_{\theta}(z)}{d z}$ is the determinant of the generative model's Jacobian matrix.

To keep the computation of the second term tractable, the neural network function, $G_{\theta}(\cdot)$, is restricted to forms such as NICE\cite{dinh2014nice}, Real-NVP\cite{dinh2016density} and Glow\cite{kingma2018glow}. In these network architectures, the Jacobian matrix is a multiplication of only lower triangular matrices or quasi-diagonal matrices, which leads to efficient computation of the determinant. 

\section{Method} \label{sec:method}
In this paper, we propose a variational Bayesian method to learn an approximate posterior distribution for the purpose of efficiently characterizing uncertainty in underdetermined imaging systems.
We parameterize the latent image distribution using a flow-based generative model,
\begin{equation} \label{eq:flow}
    x \sim q_{\theta}(x)  \quad \Leftrightarrow \quad x = G_{\theta}(z) ,z \sim \mathcal{N}(0, 1)
\end{equation}
and learn the model's weights by minimizing the Kullback–Leibler (KL) divergence between the generative model distribution, $q(x)$, and the image posterior distribution, $p(x|y) \propto p(y|x)p(x)$:
\begin{align} \label{eq:kl}
       & \theta^{\star} = \textrm{arg} \min_{\theta} D_{\textrm{KL}} (q_{\theta}(x)\|p(x|y)) \\
        \notag &= \textrm{arg} \min_{\theta} \int q_{\theta}(x) [\log q_{\theta}(x) - \log p(x|y)] dx \\
       \notag  &= \textrm{arg} \min_{\theta} \int q_{\theta}(x) [\log q_{\theta}(x) - \log p(y|x) - \log p(x)] dx \\
      \notag  &= \textrm{arg} \min_{\theta} \mathbb{E}_{x \sim q_{\theta}(x)}[- \log p(y|x) - \log p(x) + \log q_{\theta}(x)].
\end{align}
Note that this loss can be interpreted as an an expectation over the maximum {\it a posteriori} (MAP) loss from Eq.~\ref{eq:map} with an added term to encourage entropy on the image distribution. Minimizing the negative entropy term, $H_{\theta} = \mathbb{E}_{x \sim q_{\theta}(x)}[\log q_{\theta}(x)]$, prevents the generative model from collapsing to a deterministic solution.

For most deep generative models the sample likelihood, $q_{\theta}(x)$, cannot be evaluated exactly. 
However, since a sample's likelihood can be computed according to Eq.~\ref{eq:changeofvar} for flow-based models, this stochastic optimization problem can be rewritten as
\begin{align} \label{eq:dpi}
       \notag  \theta^{\star} = \textrm{arg} \min_{\theta} & \mathbb{E}_{z \sim \mathcal{N}(0, 1)} \{- \log p(y|G_{\theta}(z)) - \log p(G_{\theta}(z))\\
        & \hspace{0.3in} \left. + \log \pi(z) - \log \left| \det \frac{d G_{\theta}(z)}{d z} \right| \right\}.
\end{align}
Approximating the expectation using a Monte Carlo method, and replacing the data likelihood and prior terms with the data fitting loss and regularization functions from Eq.~\ref{eq:rml}, we obtain the optimization problem
\begin{align} \label{eq:dpimc}
     \notag   \theta^{\star} = \textrm{arg} \min_{\theta} \sum_{k=1}^N  \{ &\mathcal{L}(y, f(G_{\theta}(z_k))) + \lambda \mathcal{R}(G_{\theta}(z_k)) \\  &  \left. - \log \left| \det \frac{d G_{\theta}(z_k)}{d z_k} \right| \right\},
\end{align}
where $z_k \sim \mathcal{N}(0, 1)$, $N$ is the number of Monte Carlo samples, and the term $\log \pi(z_k)$ is omitted since its expectation is constant. The expectation of the data fitting loss and image regularization loss are optimized by sampling images from the generative model $G_{\theta}(\cdot)$. Note that when $\mathcal{L}(\cdot, \cdot)$ does not define the true data likelihood, or $\mathcal{R}(\cdot)$ does not define an image prior, the learned network only models an approximate image posterior instead of the true posterior distribution.

The data fitting loss and the regularization function are often empirically defined and may not match reality perfectly. 
Recalling that the third term is related to the entropy of the learned distribution, similar to $\beta$-VAE\cite{higgins2016beta}, we introduce another tuning parameter $\beta$ to control the diversity of the generative model samples,
\begin{align} \label{eq:dpimcbeta}
     \notag   \theta^{\star} = \textrm{arg} \min_{\theta} \sum_{k=1}^N  \{ &\mathcal{L}(y, f(G_{\theta}(z_k))) + \lambda \mathcal{R}(G_{\theta}(z_k)) \\ &  \left. - \beta \log \left| \det \frac{d G_{\theta}(z_k)}{d z_k} \right| \right\}.
\end{align}
When the uncertainty of the reconstructed images seems smaller than expected, we can increase $\beta$ to encourage higher entropy of the generative distribution; otherwise, we can reduce $\beta$ to reduce the diversity of reconstructions. Larger $\beta$ also encourages more exploration during training, which can be used to accelerate convergence.  

\begin{figure}[!t]
\centering
\setlength{\fboxrule}{0pt}
\framebox[\columnwidth]{\includegraphics[width=0.99\columnwidth]{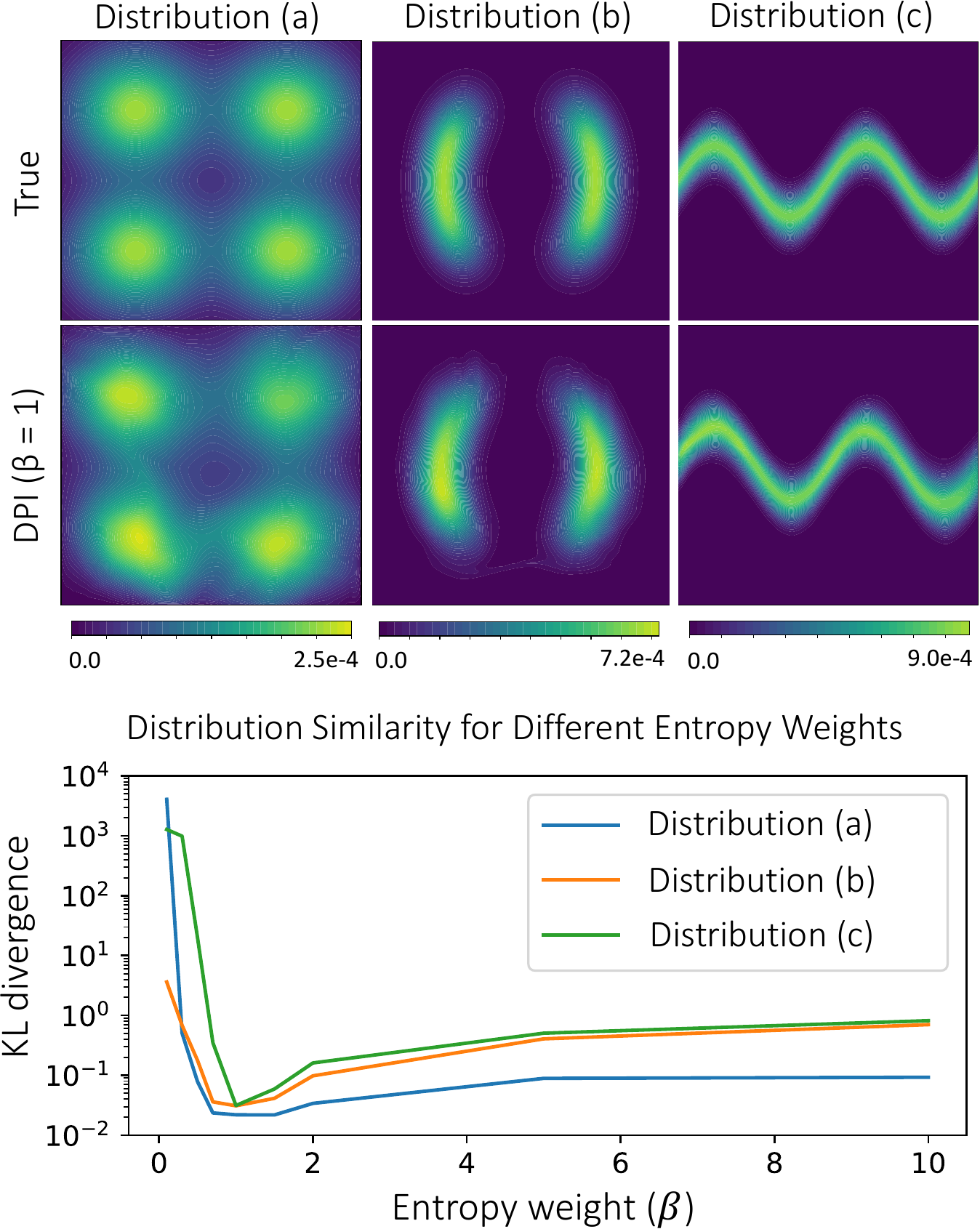}}
\vspace*{-8mm}
\caption[example2d]{(Top) Three two-dimensional posterior distributions and their learned DPI approximations for entropy weight $\beta=1$. Contours are overlaid to highlight differences between the true and learned distributions. (Bottom) Similarity between the true and learned distributions under different entropy weights $\beta$. As expected, $\beta=1$ minimizes the KL divergence between the two distributions.}\vspace{-.2in}
\label{fig:example2d}
\end{figure}

\vspace{-.1in}
\section{Toy Examples} \label{sec:toy}
\vspace{-.0in}
We first study our method using two-dimensional toy examples. 
Assuming $x$ is two-dimensional, and the joint distribution $p(y, x)$ is given exactly by the potential function $\mathcal{J}(x)$, Eq.~\ref{eq:dpimcbeta} can be simplified to
\begin{equation}
    \begin{split}
        \theta^{\star} \hspace{-.05in}&= \hspace{-.04in} \textrm{arg} \min_{\theta} \left\{ \hspace{-.05in} \mathbb{E}_{z\sim \mathcal{N}(0, 1)}  [\mathcal{J}(G_{\theta}(z))] \hspace{-.04in} - \hspace{-.04in} \beta \log \left| \det \frac{d G_{\theta}(z_k)}{d z_k} \right|\right\}.
    \end{split}
\end{equation}
For the following toy tests the generative model $G_{\theta}(\cdot)$ is designed using a Real-NVP architecture with 32 affine coupling layers~\cite{dinh2016density}. 

We test the approach on 3 joint distribution functions: (a) a Gaussian mixture potential, (b) a bowtie potential, and (c) a Sinusoidal potential. Fig.~\ref{fig:example2d} shows the true and learned probability density function in these three cases for $\beta=1$. 
Qualitatively, the learned generative model distributions match the true distributions well. 

As derived in Eq.~\ref{eq:kl}, the posterior should be best learned by $G_{\theta}(\cdot)$ for entropy weight $\beta=1$. To test this claim, we adjust $\beta$ from $0.1$ to $10$ to see how the entropy term influences the learned generative model distribution. According to the graph of KL divergence versus $\beta$ in Fig.~\ref{fig:example2d}, the learned distributions match the true distributions best when the entropy weight equals $1$. This verifies the theoretical expectation presented in the Method section.

Since the generative model is a transformation of a continuous multivariate Gaussian distribution, the learned distribution is also continuous. This leads to a common issue in flow-based generative models: there are often a few samples located in the high loss regions when the modes are not connected (see Fig.~\ref{fig:example2d} distribution (a) and (b)). Some approaches\cite{gao2020flow} have been proposed recently to solve this problem, however, in this paper we neglect this issue and leave it for future work.

\vspace{-.1in}
\section{Interferometric Imaging Case Study} \label{sec:interferometry} 

\begin{figure}[!t]
\centering
\setlength{\fboxrule}{0pt}
\framebox[\columnwidth]{\includegraphics[width=0.95\columnwidth]{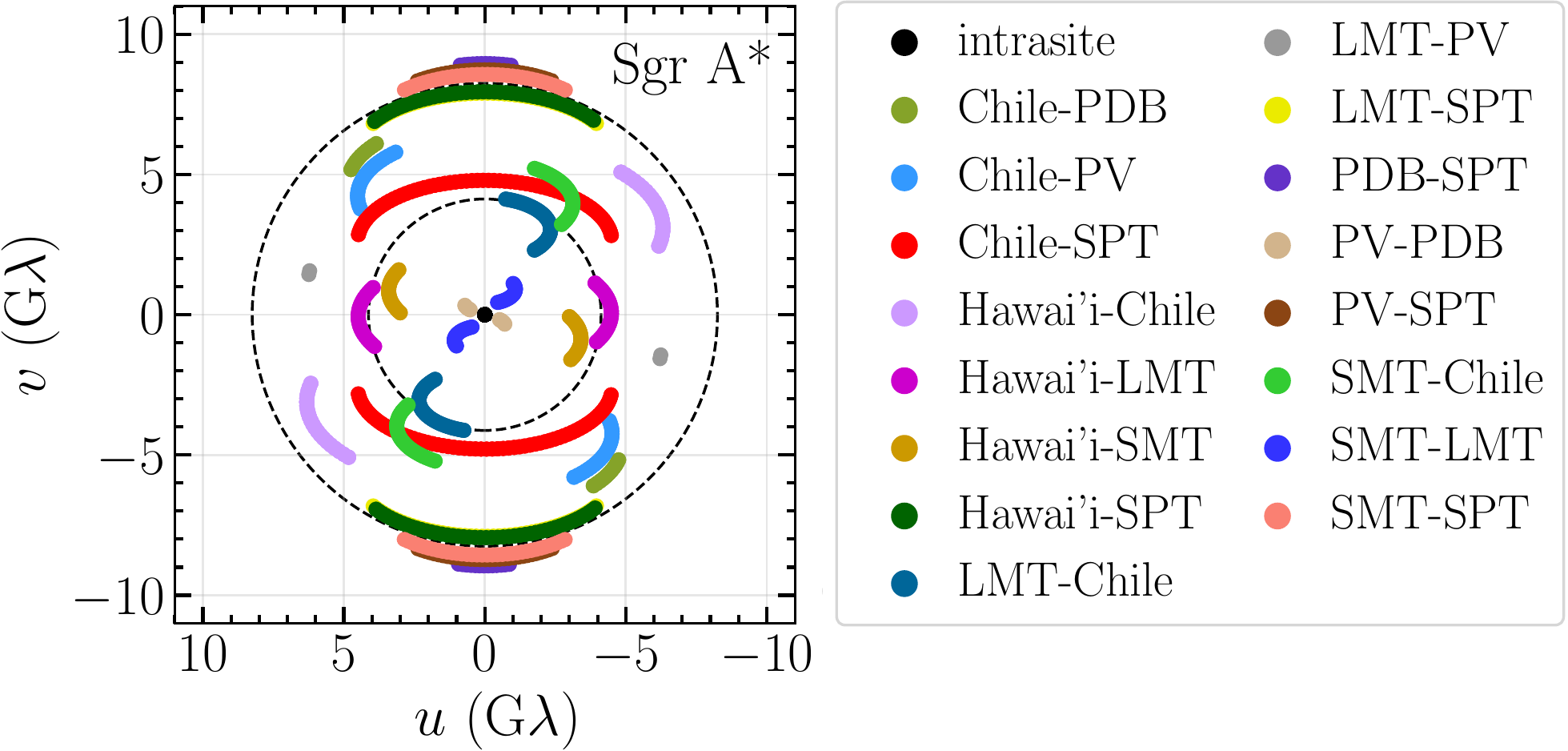}}
\vspace*{-8mm}
\caption[sgrA]{Spatial frequency samples for a 9-telescope EHT array observing the black hole Sgr A* over a night. 
The $(u,v)$ position indicates the 2D spatial Fourier component of the image that is sampled by a telescope pair.
}\vspace{-.1in}
\label{fig:sgrA}
\end{figure}

We demonstrate our proposed approach on the problem of radio interferometric astronomical imaging~\cite{richard2017interferometry}. In interferometric imaging, radio telescopes are linked to obtain sparse spatial frequency measurements of an astronomical target. These Fourier measurements are then used to recover the underlying astronomical image. Since the measurements are often very sparse and noisy, there can be multiple image modes that fit the observed data well. The Event Horizon Telescope (EHT) used this technique to take the first picture of a black hole, by linking telescopes across the globe\cite{akiyama2019first}. Fig.~\ref{fig:sgrA} shows the spatial frequency (Fourier) samples that can be measured by a 9-telescope EHT array when observing the black hole Sagittarius A* (Sgr A*)\footnote{The largest sampled spatial frequency determines the interferometer's nominal resolution of $\approx$ 25$\mu as$ for the EHT. In this paper, we neglect evolution of Sgr A* and assume it is static over the course of a night.}.

Each Fourier measurement, or so-called ``visibility", is obtained by correlating the data from a pair of telescopes. The measurement equation for each visibility is given by
\begin{equation} \label{eq:vis}
    V_{a,b}^t = g_a^t g_b^t \exp\left[-i(\phi_a^t - \phi_b^t)\right] F_{a,b}^t x + n_{a,b}^t,
\end{equation}
where $a$ and $b$ index the telescopes, $t$ represents time, and $F_{a,b}^t x$ extracts the Fourier component from image $x$ corresponding to the baseline between telescope $a$ and $b$ at time $t$. The measurement noise comes from three sources: (1) time-dependent telescope-based gain error, $g_a^t$ and $g_b^t$, (2) time-dependent telescope-based phase error, $\phi_a^t$ and $\phi_b^t$, and (3) baseline-based Gaussian thermal noise, $n_{a,b}^t \sim \mathcal{N}(0, \nu_{a, b}^2)$. The standard derivation of thermal noise depends on each telescope's System Equivalent Flux Density (SEFD):
\begin{equation} \label{eq:sefd}
    \nu_{a, b} \propto \sqrt{SEFD_{a} \times SEFD_{b}}.
\end{equation}

When the gain and phase errors are reasonably small, the interferometric imaging problem is approximately a convex inverse problem. However, when the gain errors and the phase errors (caused by atmospheric turbulence and instrument miscalibration) are large, the noisy visibilities can be combined into robust data products that are invariant to telescope-based errors, termed closure phase, $C_{a,b,c}^{\mbox{ \tiny ph.},t}$, and closure amplitude, $C_{a,b,c,d}^{\mbox{ \tiny amp.}, t}$\cite{chael2018interferometric}:
\begin{equation} \label{eq:closure}
    \begin{split}
        C_{a,b,c}^{\mbox{ \tiny ph.},t} &= \angle \left( V_{a,b}^t V_{b,c}^t V_{c,a}^t\right) \approx f_{a,b,c}^{\mbox{ \tiny ph.}, t}(x),\\
        C_{a,b,c,d}^{\mbox{ \tiny amp.}, t} &= \frac{|V_{a,b}^t| |V_{c,d}^t|}{|V_{a,c}^t| |V_{b,d}^t|} \approx f_{a,b,c,d}^{\mbox{ \tiny amp.}, t}(x).
    \end{split}
\end{equation}
These nonlinear ``closure quantities" can be used to constrain non-convex image reconstruction 
\begin{equation} \label{eq:closureloss}
\begin{split}
    \mathcal{L}&(y, f(x)) = \sum_{a, b, c} |C_{a,b,c}^{\mbox{ \tiny ph.},t} - f_{a,b,c}^{\mbox{ \tiny ph.}, t}(x)|^2 / \sigma_{a,b,c}^2 \\
    &+ \sum_{a, b, c, d} | \log C_{a,b,c, d}^{\mbox{ \tiny amp.},t} - \log f_{a,b,c,d}^{\mbox{ \tiny amp.}, t}(x)|^2 / \sigma_{a,b,c,d}^2,
\end{split}
\end{equation}
where $\sigma_{a,b,c}$ and $\sigma_{a,b,c,d}$ are the standard deviations of the corresponding closure term computed based on SEFDs. Note the closure quantities are not independent and the corresponding standard deviations are derived from linearization, so Eq.~\ref{eq:closureloss} only approximates the true data log-likelihood.

In the following contents, we demonstrate our Deep Probabilistic Imaging (DPI) approach on both convex reconstruction with complex visibilities and non-convex reconstruction with closure quantities using both synthetic and real datasets. With this new approach, we successfully quantify the uncertainty of interferometric images, as well as detect multiple modes in some data sets.

\vspace*{-2mm}
\subsubsection{Implementation}
\begin{figure}[!t]
\centering
\setlength{\fboxrule}{0pt}
\framebox[\columnwidth]{\includegraphics[width=1.0\columnwidth]{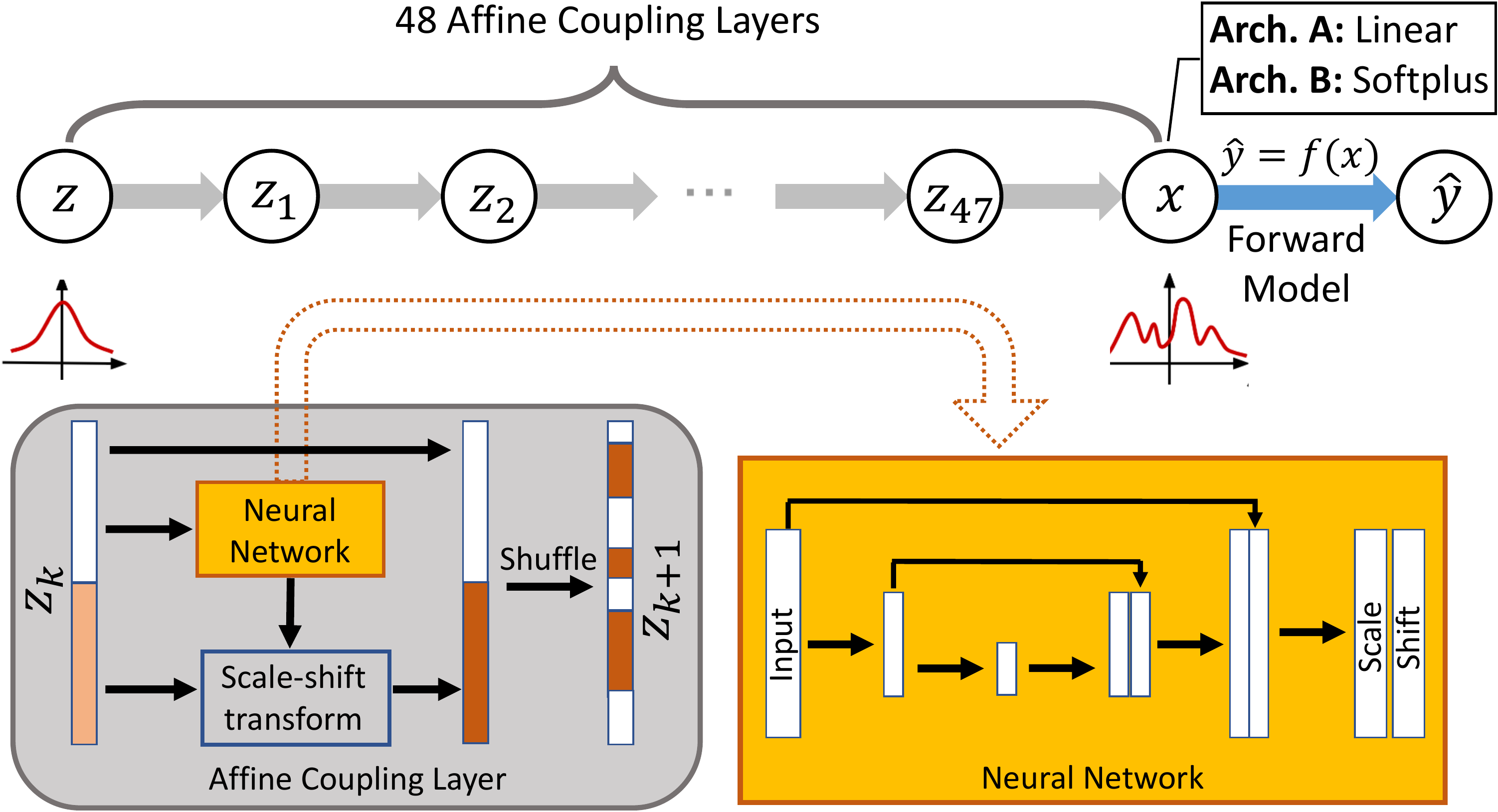}}
\vspace*{-6mm}
\caption[realnvp]{The Real-NVP network architecture used in DPI interferometric imaging. It consists of 48 affine coupling layers, transforming a latent Gaussian sample $z$ to an image reconstruction sample, $x$. In each affine coupling layer, the input vector, $z_k$, is split into two parts: the first half is kept unchanged, while the second half is modified based on a neural network transformation of the first half. The network is composed of dense layers (including the Leaky ReLU activation and the batch normalization) with skip connections, similar to a UNet. After each affine-coupling transform block, the vector is randomly shuffled, so that different elements are modified in the next transform block.
}\vspace{-.15in}
\label{fig:realnvp}
\end{figure}

For all interferometric imaging DPI results we use a Real-NVP \cite{dinh2016density} network architecture with 48 affine coupling layers (Fig.~\ref{fig:realnvp}). The scale-shift function of each layer is modeled as a U-Net style fully connected network with five hidden layers and skip connections. We train the model using Adam with a batch size of 32 for 20,000 epochs. 
We note that a limitation of our general approach is the modeling capacity of the flow-based generative model used. We find the proposed architecture is satisfactory for characterizing uncertainty in images of size 32 $\times$ 32 pixels. 

\begin{figure}[!t]
\centering
\setlength{\fboxrule}{0pt}
\framebox[\columnwidth]{\includegraphics[width=1.0\columnwidth]{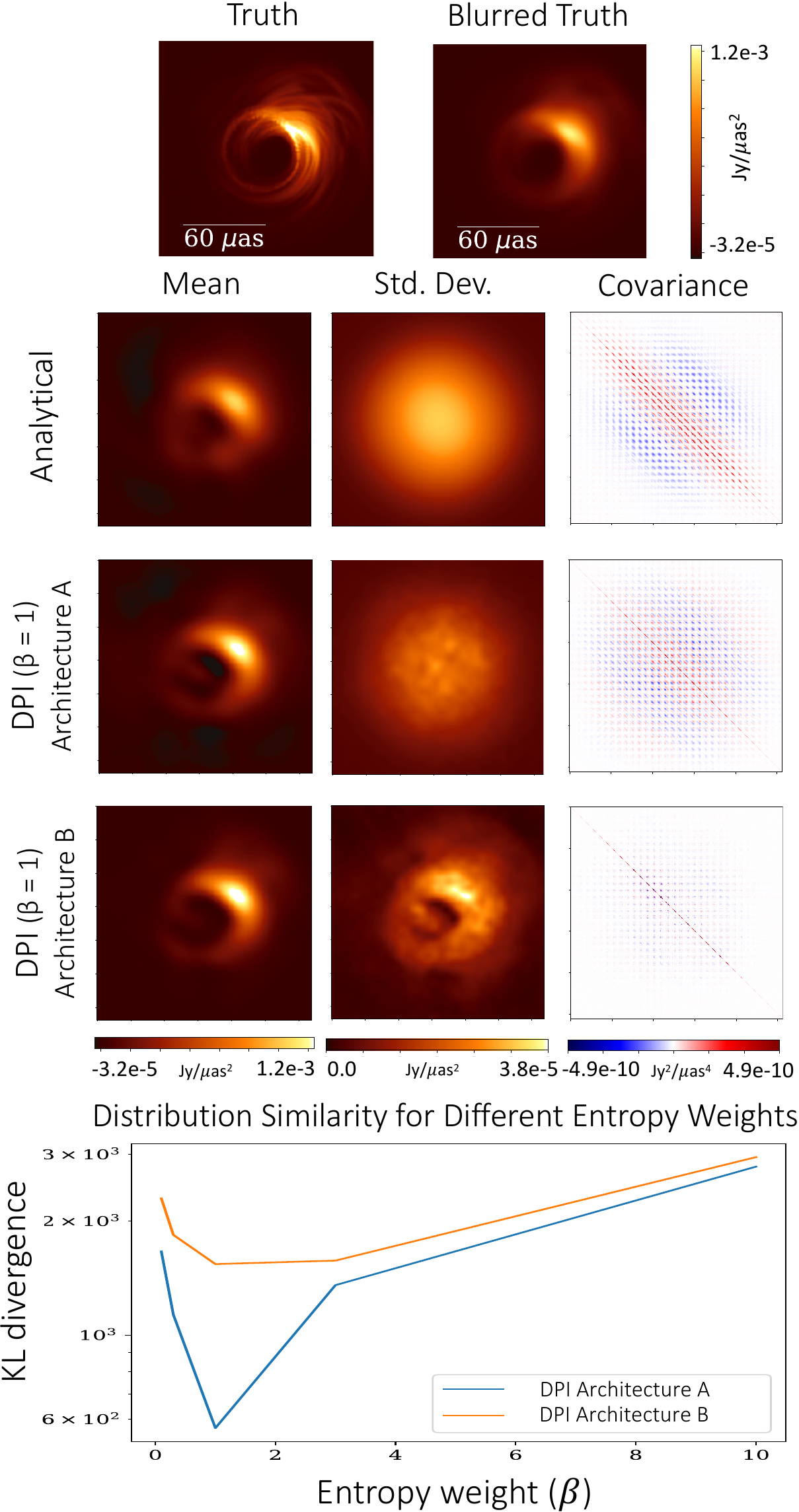}}
\caption[examplevis]{Posterior distribution estimation for convex interferometric image reconstruction. (Top) The target synthetic image of a black hole (original and blurred to 0.4$\times$ the interferometer's nominal resolution of $\approx$ 25 $\mu as$). Noisy visibility measurements are derived from this target image. (Middle) The posterior mean, pixel-wise standard deviation, and full covariance obtained analytically using a Gaussian image prior and with two DPI generative model architectures. Architecture A allows negative pixel values, while Architecture B restricts images to be non-negative. The DPI results are computed according to 2048 samples from each generative model. (Bottom) The similarity between the DPI distributions and the analytical distribution under different entropy weights $\beta$. Both architectures achieve a minimum at $\beta=1$.}
\label{fig:examplevis}
\end{figure}

\vspace*{-2mm}
\subsection{Convex Imaging with Visibilities}
In this section, we demonstrate DPI on convex interferometric imaging. In particular, the gain and phase errors in Eq.~\ref{eq:vis} are assumed to be zero (i.e., $g_a^t=g_b^t=1$ and $\phi_a^t=\phi_b^t=0$) so that complex visibilities are a linear function of the latent image. Since the thermal noise on $V_{a,b}^t$ is independent and Gaussian, we write the conditional likelihood as
\begin{equation} \label{eq:visloss}
    \mathcal{L}(y, f(x)) = \frac{1}{2}(y - F x)^T \Sigma^{-1} (y-F x),
\end{equation}
where $x$ is a column vectorized image with $M^2$ pixels,  $y = [\cdots, \Re\{V_{a, b}^t\}, \Im\{V_{a, b}^t\}, \cdots]$ is a column vector of $K$ complex visibility measurements, $F = [\cdots, \Re\{F_{a, b}^t\}, \Im\{F_{a, b}^t\}, \cdots]$ is a under-sampled Fourier transform matrix of size $K \times M^2$, and $\Sigma = \textrm{diag}([\cdots, \nu_{a, b}^2,  \nu_{a, b}^2, \cdots])$ is a $K \times K$ measurement covariance matrix derived according to the telescopes' SEFD.

In order to verify that the flow-based generative model can learn the posterior distribution of reconstructed images, we employ a multivariate Gaussian image prior,
\begin{equation} \label{eq:gaussprior}
    \mathcal{R}(x) = \frac{1}{2}(x - \mu)^T \Lambda^{-1} (x - \mu),
\end{equation}
where $\mu$ is a mean image, and $\Lambda$ is the covariance matrix defined by the empirical power spectrum of an image\cite{bouman2018reconstructing}. Since both the measurement and image prior follow Gaussian distributions, the reconstructed image's posterior distribution is also a Gaussian distribution and can be analytically derived as
\begin{align} \label{eq:analyticalsolution}
     \notag    p(x|y) \propto p(&y|x)  p(x) = \mathcal{N}_y(Fx, \Sigma) \mathcal{N}_x(\mu, \Lambda) = \mathcal{N}_x(m, C)\\
    \notag    \ m&=\mu + \Lambda F^T (\Sigma + F \Lambda F^T)^{-1} (y-F \mu)\\
        C &= \Lambda - \Lambda F^T (\Sigma + F \Lambda F^T) F \Lambda.
\end{align}


Using the specified data likelihood and prior, we train a DPI flow-based network to produce image samples of size 32$\times$32 pixels with a field of view of 160 micro-arcseconds ($\mu as$).
Fig.~\ref{fig:examplevis} demonstrates DPI on a synthetic interferometric imaging example, and compares the learned generative model distribution with the analytical posterior distribution. 
Visibility measurements are derived from the synthetic black hole image shown in the top of Fig.~\ref{fig:examplevis} (with a total flux of 2 Janskys).
The second row of the Fig.~\ref{fig:examplevis} shows the analytic posterior's mean, standard deviation, and full covariance. 
The third and fourth rows of the figure show the mean, standard deviation, and full covariance empirically derived from DPI samples under two slightly different $G_{\theta}(\cdot)$ architectures: (A) the third row uses a model with 48 affine coupling layers, and (B) the fourth row adds an additional Softplus layer to the model to enforce the physical constraint of non-negativity in the image distribution.
Without a non-negativity constraint, Architecture A's learned distribution is very similar to the analytical posterior distribution, since it is solving a same Bayesian estimation problem as defined in Eq.~\ref{eq:analyticalsolution}. 
However, this Bayesian estimation problem does not constrain the image to be positive; as a result, the central depression in the image has an inflated standard deviation. 
Architecture B's non-negative model results in a more constraining uncertainty map while achieving a slightly higher resolution reconstruction. This example also demonstrates how DPI can introduce implicit regularization through the neural network architecture.

Image distributions with different levels of sample diversity can be learned by adjusting the entropy loss weight, $\beta$. 
As expected, both generative models reach lowest KL divergence with the analytic distribution when $\beta=1$.


\begin{figure}[!t]
\centering
\setlength{\fboxrule}{0pt}
\framebox[\columnwidth]{\includegraphics[width=0.935\columnwidth]{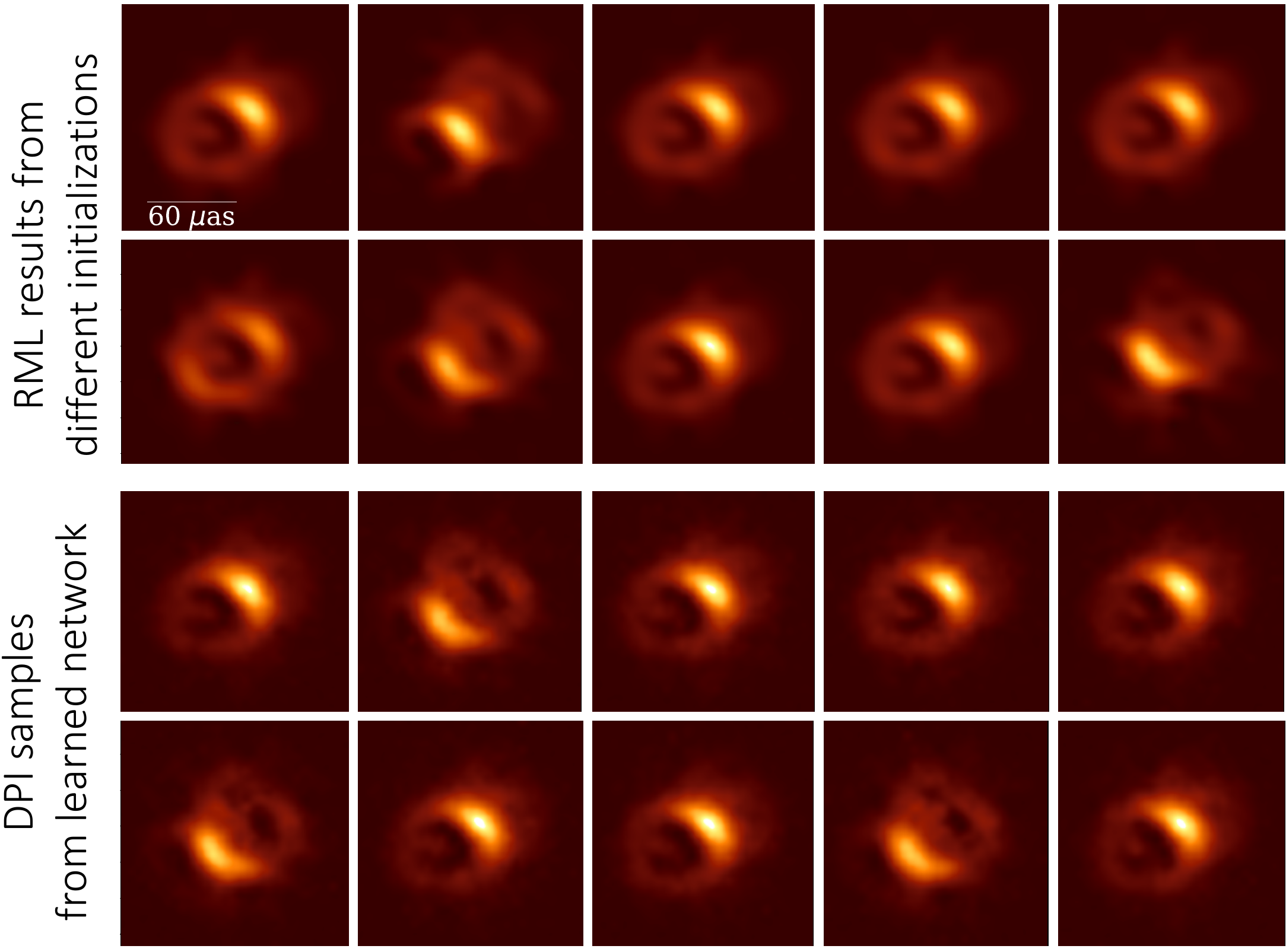}}
\vspace*{-8mm}
\caption[rmlflow]{Non-convex interferometric imaging results with closure quantities. (Top) RML reconstructed images obtained from different initializations. (Bottom) Samples from a learned DPI flow-based generative model.}\vspace{-.1in}
\label{fig:rmlflow}
\end{figure}

\begin{figure}[!t]
\centering
\setlength{\fboxrule}{0pt}
\framebox[\columnwidth]{\includegraphics[width=0.95\columnwidth]{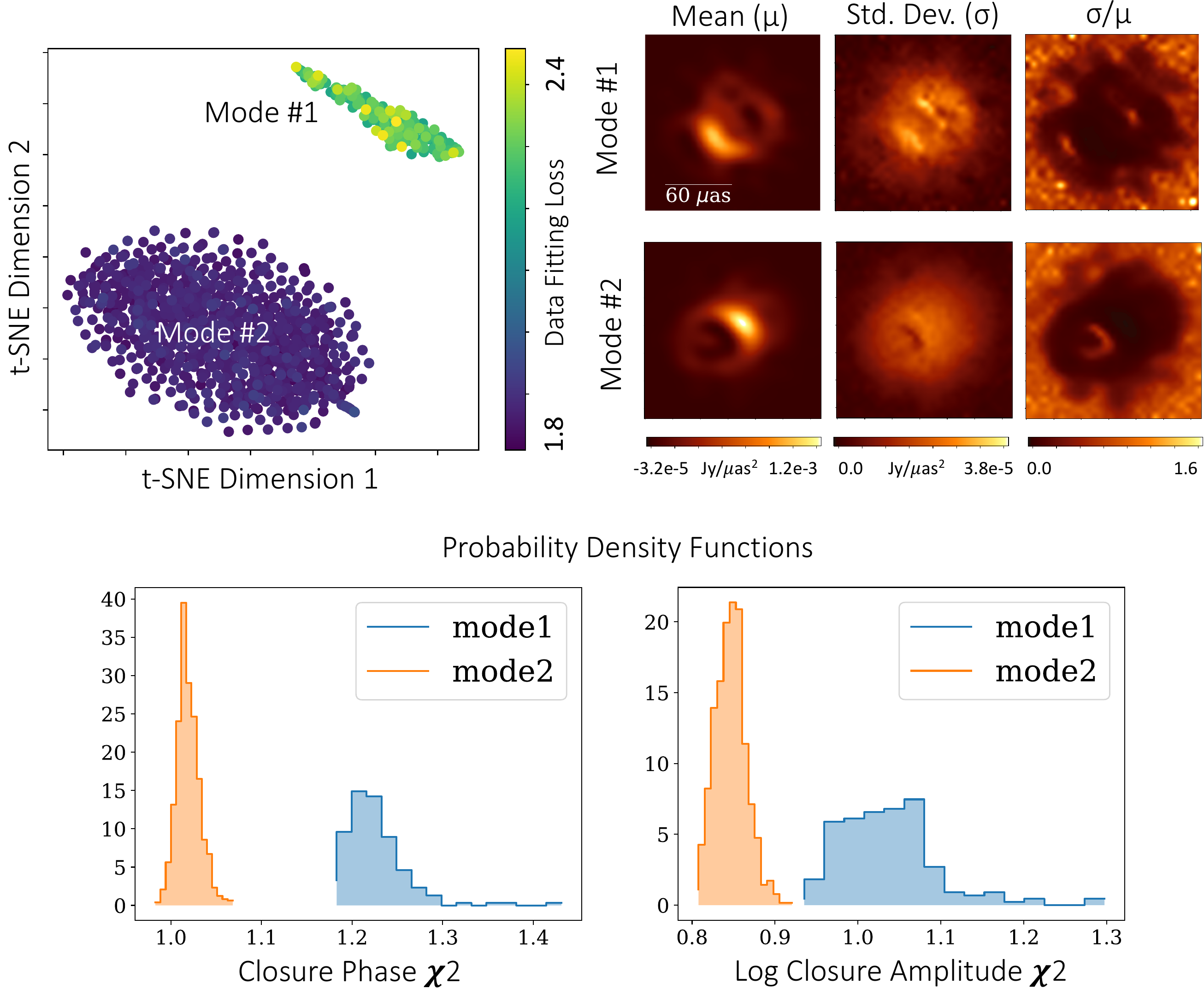}}
\vspace*{-8mm}
\caption{Analysis of 1024 reconstructed images sampled from a DPI generative model trained with closure quantities. (Top Left) two-dimensional t-SNE plot of samples with perplexity$=20$. The samples clearly cluster into two modes. (Top Right) The mean, standard deviation, and fractional standard deviation for samples from each mode. (Bottom) The distributions of data fitting losses (reduced $\chi^2$) of samples from each mode. A $\chi^2$ value of 1 is optimal for high SNR data. The second mode, which happens to be the correct solution, results in a distribution with smaller data fitting losses.}\vspace{-.25in}
\label{fig:twomodes}
\end{figure}


\vspace*{-2mm}
\subsection{Non-convex Imaging with Closure Quantities} 
In this section, we demonstrate DPI on non-convex interferometric imaging, where we reconstruct images using the closure quantities defined in Eq.~\ref{eq:closure}. With this non-convex forward model, the posterior of reconstructed images cannot be analytically computed, but it can be estimated using DPI. In all DPI reconstructions, the resulting images are 32$\times$32 pixels and result from the non-negative Real-NVP model discussed above (Architecture B). 

\vspace*{-1mm}
\subsubsection{Multi-modal Posterior Distributions}

A serious challenge for non-convex image reconstruction is the potential for multi-modal posterior distributions: visually-different solutions 
fit the measurement data reasonably well. 
In some cases, multiple modes can be identified by optimizing a regularized maximum likelihood (RML) loss with different image initializations; for example, Fig.~\ref{fig:rmlflow} (top) shows ten RML reconstructed images obtained using the closure quantities (Eq.~\ref{eq:closureloss}) from the target shown in Fig.~\ref{fig:examplevis} and the multivariate Gaussian regularizer defined in Eq.~\ref{eq:gaussprior}. 
From these results two potential image modes, which appear to be roughly 180 degree rotations of one another, clearly stand out as fitting the data well. 

\begin{figure*}[!t]
\centering
\setlength{\fboxrule}{0pt}
\framebox[\textwidth]{\includegraphics[width=0.83\textwidth]{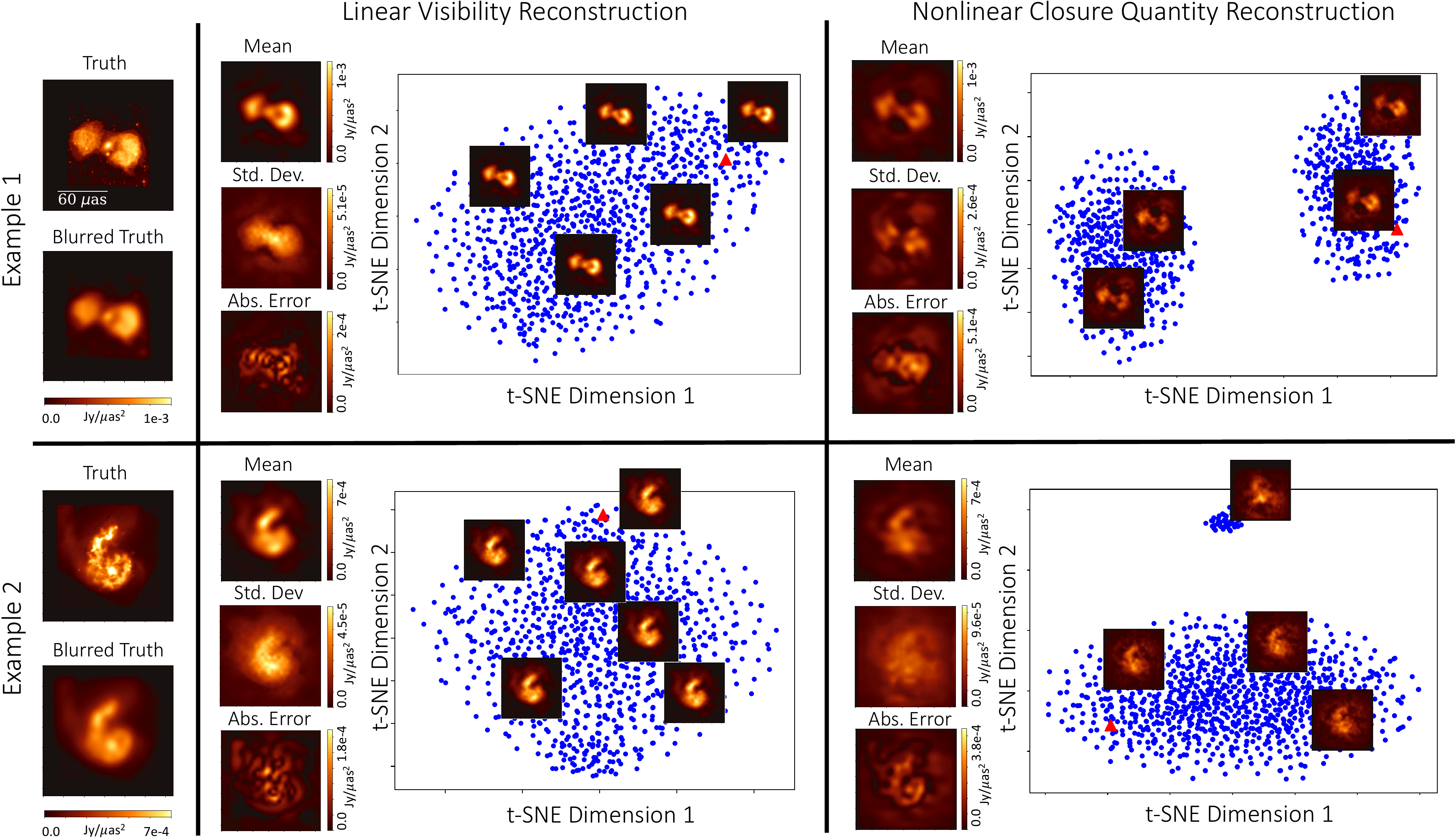}}
\vspace*{-6mm}
\caption{Examples of uncertainty quantification in convex and non-convex interferometric imaging with multivariate Gaussian regularization. In each case, the mean, standard deviation, and absolute error between the reconstruction mean and blurred truth are reported. t-SNE plots (perplexity$=20$) are used to visualize the distributions of 1024 image samples in a two-dimensional embedded space. Within the t-SNE plots, each small image corresponds to a sample embedded at its bottom-left corner. In convex interferometric imaging the area of high error approximately matches the uncertainty estimated by DPI. In non-convex interferometric imaging, both examples produced two solution modes.
The red triangle marks the embedding of the blurred truth image, which appears close to samples in the embedded image space.
}
\vspace{-.15in}
\label{fig:visualize}
\end{figure*}

\begin{figure}[!t]
\centering
\setlength{\fboxrule}{0pt}
\framebox[\columnwidth]{\includegraphics[width=1.0\columnwidth]{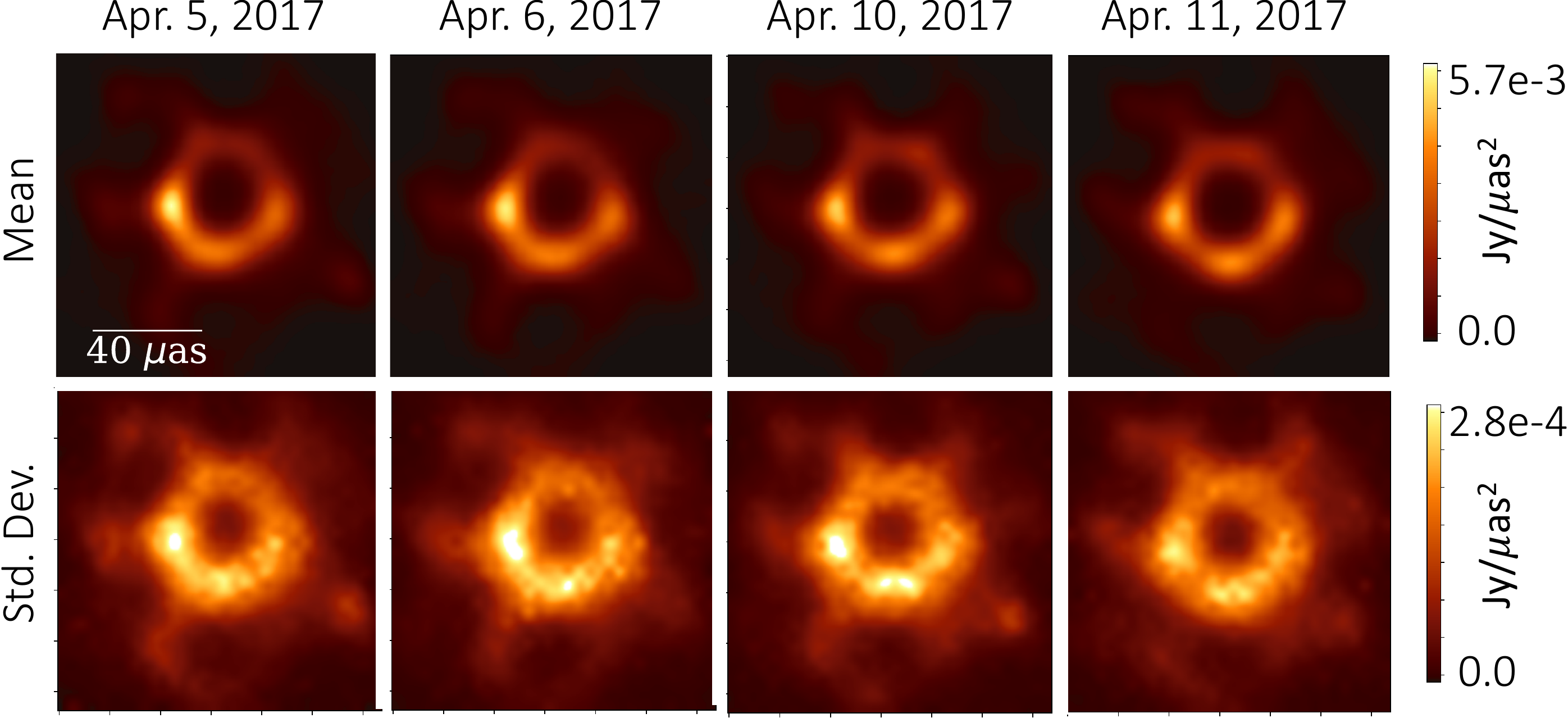}}
\vspace*{-6mm}
\caption[realdata]{DPI imaging results using real EHT 2017 observation data of the black hole in M87 on different days. See \cite{akiyama2019first} for the $(u,v)$ spatial frequency coverage for each day.} 
\vspace{-.25in}
\label{fig:realdata}
\end{figure}

Fig.~\ref{fig:rmlflow} (bottom) shows ten images sampled from a DPI flow-based generative model learned with multivariate Gaussian regularization. 
Note that the single generative model has captured the two modes identified by multiple runs of RML. 

Fig.~\ref{fig:twomodes} analyzes 1024 generative samples from a DPI model learned with multivariate Gaussian regularization. The dimensionality reduction t-SNE plot\cite{maaten2008visualizing} indicates a clustering of samples into two modes. Figure~\ref{fig:twomodes} (top right) shows the pixel-wise mean, standard deviation and fractional standard deviation of samples for each mode. 
The distributions of data fitting loss (reduced $\chi^2$) for images in each mode are shown for both closure phase and log closure amplitude constraints; a reduced $\chi^2$ value of 1 is optimal. Although it can be difficult to tell which image is correct by inspecting the statistics of a single image, by analyzing the histogram of statistics for each mode it becomes clearer which mode corresponds with the true image. 
In the supplemental material\footnote{\label{supplement}{ \scriptsize \hspace{-.04in} \url{http://imaging.cms.caltech.edu/dpi/DPIsupplement.pdf}}} we show how the resulting posterior changes under different imposed regularization.

\vspace*{-2mm}
\subsubsection{Real Interferometric Data}
In Fig.~\ref{fig:realdata} we demonstrate the performance of DPI using the publicly available EHT 2017 data, which was used to produce the first image of the black hole in M87.
In accordance with \cite{akiyama2019first}, we use a data fitting loss with not only closure quantities (Eq.~\ref{eq:closureloss}) but also roughly-calibrated visibility amplitudes. We pair this data likelihood with a combined maximum entropy (MEM) and total squared variation (TSV) regularizer (see the supplementary material for details\footnotemark[2]). Fig.~\ref{fig:realdata} shows the DPI reconstruction mean and standard deviation of M87 on different observation days. Although ground truth images are unavailable, the ring size and evolution recovered using DPI matches that seen in the original EHT imaging results. The DPI results also quantify larger uncertainty in ``knot" regions along the lower portion of the ring.


\vspace*{-2mm}
\subsection{Visualizing Uncertainty}
\vspace{-.03in}
DPI sampling provides a means to visualizing uncertainty, especially in cases where closed form approximations are insufficient. By embedding samples from our DPI model in a two-dimensional space, we are able to visualize the posterior's approximate structure. Fig.~\ref{fig:visualize} plots the embedding of DPI samples obtained using t-SNE, and shows the images corresponding to some of these samples. By plotting the blurred truth image in the same embedding we see that the truth often lies close to the posterior samples in the embedded space, though it is not guaranteed. The posterior is sensitive to the choice of image prior, and is most informative when the true image is a sample from the imposed prior.

The mean and standard deviation of DPI samples, as well as the average absolute error with respect to the blurred truth, are shown for each learned distribution in Fig.~\ref{fig:visualize}. 
Since closure quantities do not constrain the absolute flux or position of the reconstructed source, we first normalize and align samples from the closure-constrained DPI model
to account for scaled and shifted copies of similar images. 
Note that the pixel-wise standard deviation aligns well with areas of high error in the generative model's samples. 




\section{Compressed Sensing MRI Case Study} \label{sec:MRI}
Under-sampled measurements in compressed sensing MRI also result in image reconstruction uncertainty. Similar to convex interferometric imaging with visibilities, compressed sensing MRI is often modeled as a linear forward model, $y=F x + \epsilon$, where the measurements $y$ are the under-sampled $\kappa$-space spatial frequency components of the image $x$ with additive noise $\epsilon$. In this section, we apply DPI to compressed sensing MRI data with different acceleration speed-up factors and compare the DPI identified uncertainty map to the image reconstruction errors.

\begin{figure}[!t]
\centering
\setlength{\fboxrule}{0pt}
\framebox[\columnwidth]{\includegraphics[width=0.96\columnwidth]{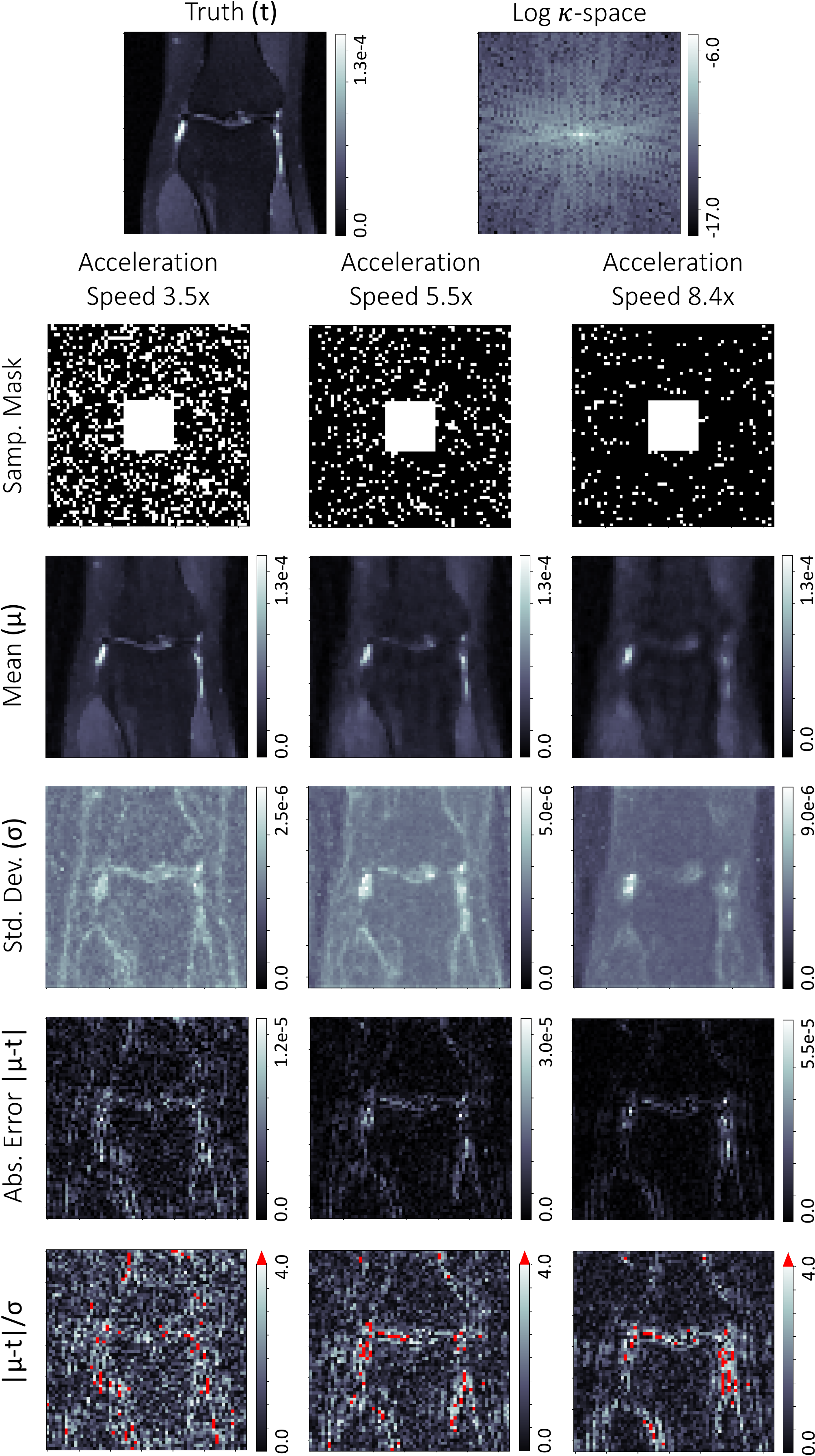}}
\vspace*{-4mm}
\caption[mri]{DPI imaging results of a synthetic compressed sensing MRI example. A knee image from the fastMRI dataset \cite{zbontar2018fastmri} is tested at three acceleration speed-up factors: $3.5\times$, $5.5\times$ and $8.4\times$ (each shown in a column). White pixels in the sampling masks (second row) indicate the observed $\kappa$-space components, where the DC (zero-frequency) component is aligned with the center of the mask. According to the pixel-wise statistics of the estimated posterior distributions (rows three and four), DPI well identifies the highly uncertain areas in the reconstructed images.}\vspace{-.25in}
\label{fig:mri}
\end{figure}

Figure~\ref{fig:mri} shows the pixel-wise statistics of the DPI estimated posterior (computed from 1024 generated images) of a synthetic example (a knee image from fastMRI dataset \cite{zbontar2018fastmri} resized to 64 $\times$ 64 pixels). The $\kappa$-space measurement noise is assumed Gaussian with a standard deviation of $0.04\%$ the DC (zero-frequency) amplitude. DPI is tested at three different acceleration speed-up factors: $3.5\times$, $5.5\times$ and $8.4\times$, i.e. 1/3.5, 1/5.5 or 1/8.4 of all $\kappa$-space components are observed. A Real-NVP network architecture similar to Fig.~\ref{fig:realnvp} with 32 affine coupling layers is used to approximate the image posterior, and a total variation (TV) regularizer is applied as the image prior. As expected, the pixel-wise standard deviation of DPI reconstruction becomes larger as the acceleration speed increases. The reconstruction values of most pixels lie within four standard deviations ($\sigma$) from the truth ($98.2\%$ pixels for $3.5\times$, $98.6\%$ pixels for $5.5\times$, and , $98.7\%$ pixels for $8.4\times$ are within $4\sigma$). Since the image posterior distribution is not pixel-wise independent Gaussian, the learned reconstruction error should not necessarily obey properties of Gaussian statistics. The profile of DPI standard deviation estimation roughly captures the pattern of the absolute reconstruction error in all cases.


\section{Conclusion} \label{sec:conclusion}
\vspace{-.0in}
In this paper, we present deep probabilistic imaging (DPI): a new framework for uncertainty quantification and multi-modal solution characterization for underdetermined image reconstruction problems. The method parameterizes the posterior distribution of the reconstructed image as an untrained flow-based generative model, and learns the neural network weights using a loss that incorporates the conditional data likelihood, prior of image samples, and the model's distribution entropy.  

We demonstrate the proposed method on toy examples, synthetic and real interferometric imaging problems, as well as a synthetic compressed sensing MRI problem. Experiments show the proposed method can approximately learn the image posterior distribution in both convex and non-convex inverse problems, which enables efficiently quantifying the uncertainty of reconstructed images and detecting multi-modal solutions. Code is available at \url{http://imaging.cms.caltech.edu/dpi/}.

\section{Acknowledgments}
This work was supported by NSF award 1935980: “Next Generation Event Horizon Telescope Design,” and Beyond Limits. The  authors  would  also like  to  thank  Joe Marino,  Dominic Pesce,  S. Kevin Zhou,  and  Tianwei Yin for the helpful discussions.
\bibliography{ref}
\bibliographystyle{aaai}

\end{document}